\documentclass[11pt,a4paper]{article}
\usepackage[hyperref]{emnlp2020}
\usepackage{times}
\usepackage{latexsym}
\usepackage{graphicx}
\usepackage{booktabs}       
\usepackage{adjustbox}      
\usepackage{multirow}

\usepackage{microtype}

\aclfinalcopy 
\def\aclpaperid{2463} 

\title{Fantastic Features and Where to Find Them: Detecting Cognitive Impairment with a Subsequence Classification Guided Approach}

\author{Benjamin Eyre \\
   Winterlight Labs \\
   Toronto, Canada \\
   \And
   Aparna Balagopalan \\
   Winterlight Labs \\
   Toronto, Canada \\
   \texttt{\{benjamin, aparna, jekaterina\}@winterlightlabs.com} \\
   \And
   Jekaterina Novikova \\
   Winterlight Labs \\
   Toronto, Canada \\
   }

\date{}

\begin{document}
\maketitle
\begin{abstract}
    Despite the widely reported success of embedding-based machine learning methods on natural language processing tasks
    , the use of more easily interpreted engineered features remains common in fields such as cognitive impairment (CI) detection
    . Manually engineering features from noisy text is time and resource consuming, 
    and can potentially result in features that do not enhance model performance. To combat this, we describe a new approach to feature engineering that leverages sequential machine learning models and domain knowledge to predict which features 
    help enhance performance. We provide a concrete example of this method 
    on a standard data set of CI speech and demonstrate that CI 
    classification accuracy improves by 2.3\% over a strong baseline when using features produced by this method. This demonstration provides an example of how this method can be used to assist classification in fields where interpretability is important, such as 
    health care. 
\end{abstract}

\section{
Introduction}
In recent years, word and sentence embedding-based 
methods have had a significant impact on the field of 
NLP \cite{devlin2019bert, mikolov2013efficient, pennington-etal-2014-glove, di2019enriching}. These 
approaches stand as an alternative to classical feature engineering approaches, where carefully crafted features, such as word length or part of speech tag, are extracted from text and used as input. 
Despite the promise of embedding-based methods, there are still several advantages to feature engineering. 
Most notably, using embeddings as input can lead to issues with interpretability \cite{heimerl2018interactive, hooker2019benchmark, kindermans2017reliability}, which is especially important in a healthcare domain \cite{balagopalan2020}. 
Meanwhile, feature engineering approaches directly lend themselves to easily interpretable models \cite{ribeiro2016should}. 
As such, 
feature engineering remains an important practice for fields such as health care, where interpretability is imperative. An extensive body of work has been produced where ML methods and engineered features have been applied to cognitive impairment (CI) detection~\cite{balagopalan2018effect,karlekar2018detecting,zhu2019detecting}. 

In this work, we present a new feature engineering method 
that is 
guided by 
classifying subsets of a pause-centred speech sequence (subsequences), and
inspired by literature suggesting that CI could be indicated by the words that subjects pause before \cite{calley2010subjective, mack2013word, seifart2018nouns}. This approach aims to extract pause-related information while minimizing the noise added from unrelated factors. 
This method 
generates interpretable and effective features, potentially saving time and resources spent on excess feature engineering. 
We validate this method by presenting a 2.3\% accuracy increase over a strong baseline on CI vs healthy (HC) classification, matching the state of the art \cite{hernandez2018computer}.

In summary, our major contributions are:

$\bullet$ A method of classifying speech using only a token of interest and a small context around it, i.e. \emph{subsequence classification} (Sec. \ref{sec:subsequence-classification}).

$\bullet$ A novel \emph{feature engineering approach} guided by subseqence classification (Sec. \ref{sec:our-technique}).

$\bullet$ Validating this approach by showing that it aids in achieving classification results comparable to the state of the art (Sec. \ref{sec:transcript-results}).

\section{Related work}
Several authors report increases in CI detection performance by extracting acoustic features such as filled and unfilled pause counts, as well as average pause duration \cite{toth2015automatic,toth2018speech,pistono2016pauses}. However, we believe further performance increases can be achieved if we focus not only on the pauses themselves, but also the linguistic context in which the pauses occur in speech.

Recently, \citet{hernandez2018computer} achieved an accuracy of 78\% when classifying Dementiabank (Sec. \ref{sec:DB}) transcripts as CI or HC using an extended set of lexical features, which is to the best of our knowledge the state of the art (SOTA). We 
use their recorded performance as a benchmark when validating our approach. 

Several authors have reported performance gains by using subsequences to aid with classification. 
These authors use subsequences only as a means to process full sequences \cite{phan2017audio}, or they use the presence of common subsequences as a feature for longer text sequences \cite{iglesias2007sequence, kumar2005using}. To the best of our knowledge, no prior work describes using subsequence classification to guide feature engineering.

\begin{table}
  \centering
  \begin{adjustbox}{max width=0.7\linewidth}
  \begin{tabular}{ccccc}
    \toprule
    Data Subset     & HC     & CI & Total \\
    \hline \hline
    DB (transcripts) & 229 (42\%) &321 (58\%)& 550\\
          \midrule
    DB-C1 & 317 (33\%) &645 (67\%)& 962\\
     DB-C2 & 511 (35\%) &963 (65\%)&  1,474\\
     DB-C3 &529 (35\%)& 980 (65\%)&  1,509\\
     DB-Utt   &755 (42\%)& 1,059 (58\%)&  1,814\\
    
     \bottomrule
  \end{tabular}
  \end{adjustbox}
  \caption {Overview of the number of samples (subsequences or transcripts) in different subsets of DB.}\label{tab:sample-table}
\end{table}

\section{Experimental Method}
 \label{sec:method}
In this section, we define the data sets and methodology used 
in our experimental framework.

\subsection{Data Sets}
\label{sec:DB}
 
\textbf{Dementiabank (DB):} Dementiabank\footnote{https://dementia.talkbank.org} is a large public data set of pathological speech~\cite{dembank}, containing audio files and transcripts of participants describing the ‘Cookie Theft’ image. Transcripts are created manually by trained transcriptionists following the CHAT protocol \cite{macwhinney2014childes}. 
Out of the 286 participants, 193 are diagnosed with some form of CI (CI; N = 321 transcripts) and 93 are healthy controls (HC; N = 229 transcripts). Transcripts receive a CI or HC label corresponding to whether the participant who produced the transcribed speech was cognitively impaired or not. 

\noindent\textbf{Subsequence-based Data Subsets:}
\label{sec:subsequence-extraction}
To conduct subsequence classification, we extract subsequences of varying length from each transcript. 

\begin{figure}
\includegraphics[width=72mm]{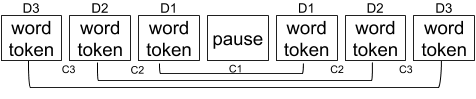}
\caption{Visualization of the difference between contexts and distances in a pause-focussed subsequence.}\label{fig:context}
\end{figure}

For each transcript in DB, 
each utterance containing a pause was extracted and labelled as positive if the sample contains CI speech, or negative otherwise. 
Subsequences were extracted from these utterances by taking the first one, two, or three speech tokens before and after each pause.\footnote{If there were less than two or three tokens before or after a pause, the largest possible sequence of tokens was extracted.} We created three data subsets by including subsequences of at most one, two, or three tokens around the pause: Context~1 (DB-C1), Context 2 (DB-C2), and Context 3 (DB-C3), respectively. We also included one data subset including full utterances that include pauses, DB-Utt (Tab.\ref{tab:sample-table}). Identical subsequences found in both classes were removed. Furthermore, subsequences extracted from HC transcripts are labeled as HC, and subsequences extracted from CI transcripts are labeled as CI. 
We refer to the tokens that are next to the pause as Distance 1 (D1), the tokens that are one token away from the pause as Distance 2 (D2), and the tokens that are two tokens away from the pause as Distance 3 (D3). The differences between \textit{context} and \textit{distance} are shown in Fig.\ref{fig:context}. For example, for the pause sequence \textit{``The boy is *uh* stealing a cookie"}, only the tokens \textit{``boy"} and \textit{``a"} would be considered the Distance 2 tokens for this sequence, while the tokens \textit{``boy"}, \textit{``is"}, \textit{``*uh*"}, \textit{``stealing"}, and \textit{``a"} would be considered the Context 2 tokens for this sequence.

\subsection{Feature Extraction}
In this section, we describe how features are extracted on the transcript-level (for transcript classification) and on the token-level (for subsequence classification). 

\noindent\textbf{Transcript-Level Features:} 
\label{sec:transcript-features}
We extract over 500 
linguistic and acoustic features from each transcript, such as part of speech counts and average word length (App.\ref{sec:app-feature-list}).
These features, referred to as the \textit{Original} feature set, are used to provide a baseline to benchmark transcript-level classification performance. 
We also use the \textit{Original} feature set as a base that we extend with newly engineered transcript-level features (Sec. \ref{sec:our-technique}). To produce an additional baseline, we perform feature selection on the \textit{Original} feature set, and found $k=85$ features led to the greatest performance.


\noindent\textbf{Token-Level Features:} 
\label{sec:token-features}
In order to conduct subsequence classification, we extract features on the token-level for each of the subsequence-based data subsets. 
Of the 
\textit{Original} feature set, we select a subset of features that have a clear token-level analogue (App.\ref{sec:app-feature-list}). For instance, the transcript-level feature of average word length has the token-level analogue of individual word length. 
After feature extraction, each token is represented by a 23-dimensional input vector. Consequently, each subsequence in the DB-C1, DB-C2, DB-C3, and DB-Utt data subsets is represented as a \emph{T} by 23 matrix, where \emph{T} is the length of the sequence in tokens. 

\subsection{Classification}
In this section, we describe the methodology used for subsequence and transcript classification.
Subsequence classification is used to guide the engineering of new features, while transcript classification validates the new features' effectiveness.

\noindent\textbf{Subsequence Classification:}
\label{sec:subsequence-classification}
Our subsequence classification experiment involves performing 5-fold cross validation with each of the DB-C1, DB-C2, DB-C3, and DB-Utt data subsets. 
Subsequences are classified as either HC or CI. We conduct classification using  GRU-based \cite{cho2014learning} models with an attention mechanism designed for document classification \cite{yang2016attention}, with model parameters tuned for each of the data subsets. We report accuracy for M-C1, the model that achieved the highest accuracy on DB-C1, M-C2, the model that achieved the highest accuracy on DB-C2, M-C3, the model that achieved the highest accuracy on DB-C3, and M-Utt, the model that achieved the highest accuracy on DB-Utt 
(additional training details provided in App.\ref{sec:appendix-models}).

\noindent\textbf{Transcript Classification:}
\label{sec:transcript-classification}
We evaluate the efficacy of our feature engineering approach (Sec.\ref{sec:our-technique}) by performing transcript-level 10-fold cross validation with a variety of feature sets. Transcripts are classified as either HC or CI. We use the \textit{Original} feature set, as well as the top 85 of the  \textit{Original} features, based on their ANOVA F-values, as baselines. Additionally, we extend the \textit{Original} feature set with the $k$ best features, based on ANOVA F-values, from each of the feature sets generated using our novel feature engineering approach (Sec. \ref{sec:our-technique}), separately. 
$k$ is optimized for accuracy for each extending feature set separately.
To classify DB transcripts, we use 5 ML models: an SVM, a gradient boosting ensemble, a 2-layer neural network (NN), a random forest, and an ensemble of the previous four models (Ens). We report the accuracy (Acc), precision (Prec), sensitivity (Sens), and specificity (Spec) for the model that achieved the greatest cross validated accuracy for each feature set, separately (training details provided in App.\ref{sec:appendix-models}).


\section{Proposed Feature Engineering Approach}
\label{sec:our-technique}
Our approach to engineering new transcript-level features involves three major steps:

\noindent\textbf{1)} Subsequences of varying length centred around a token of interest, in our case a pause, must be extracted from each of the input transcripts and grouped into subsets based on maximum length. Each of the tokens in these subsequences must have token-level features extracted. The token-level features, as well as the central token, should be chosen based on in-domain knowledge. 

\noindent\textbf{2)} A sequential ML model must be cross validated on each of the subsequence data subsets from the previous step in a subsequence classification experiment. 
Here, we are specifically attempting to exploit the ability for sequential machine learning models to uncover patterns in sequential data. The mean cross validated accuracy on each of these length-based data subsets should be used as an indicator of how much distinguishing information can be extracted from tokens within the specified range of the pause.

\noindent\textbf{3)} Based on the recorded cross validated accuracies from the previous step, 
transcript-level aggregations of the token-level features must be created at various distances from the pause. 
We propose two methods of aggregating token-level features (DB-specific details provided in App.\ref{sec:app-feature-list}):

$\bullet$\textbf{ Continuous features} can be aggregated simply by taking the average of a feature across each of the tokens. An example of this would be calculating the average word length for each of the tokens found at a specified distance from a pause.

$\bullet$\textbf{ Categorical features} can be aggregated using counts or ratios, such as the number of nouns occurring at a specified distance from a pause.

\begin{table*}
  \centering
  \begin{adjustbox}{max width=0.6\linewidth}
  \begin{tabular}{cccccc}
    \toprule
    Feature Set & Model & Acc & Prec & Sens & Spec \\
    \hline \hline
     Original & Ens &74.77$\pm$0.6* & 82.08$\pm$0.4* & 73.1$\pm$0.5* & 79.74$\pm$0.4* \\
     Original w/ feat.sel. & Ens & 75.18$\pm$1.4 &83.37$\pm$1.2  & 72.21$\pm$1.3*   & 81.67$\pm$1.6   \\
     \midrule
     Original + F-D1 & NN & 74.41$\pm$1.9* & 78.59$\pm$1.1*& \textbf{77.15$\pm$3.3} & 72.63$\pm$1.9* \\
     Original + F-D2 & Ens & \textbf{77.09$\pm$1.0} & \textbf{84.40$\pm$0.8}& 75.21$\pm$1.0 & 82.32$\pm$0.9 \\
     Original + F-D3 & Ens &76.05$\pm$0.8 & 84.02$\pm$0.9& 71.92$\pm$0.7* & \textbf{83.33$\pm$1.3} \\
     Original + F-C2 & NN &75.14$\pm$1.4 & 79.85$\pm$1.5*& 76.93$\pm$0.9* & 74.02$\pm$2.5* \\
     Original + F-C3 & Ens & 74.82$\pm$1.2* & 83.68$\pm$1.2& 70.62$\pm$1.9* & 82.63$\pm$1.4 \\
     \bottomrule
  \end{tabular}
  \end{adjustbox}
  \caption {Transcript classification performance for each feature set's best performing classification model, averaged across four random seeds. Bold indicates best performance, and * indicates significance ($p<0.05$) when compared to the model using F-D2 features.}\label{tab:transcript-results}
\end{table*}

These transcript-level aggregates should only be extracted for the distances that produced the greatest cross validated accuracy during subsequence classification, as the subsequence classification performance indicates that the features found in that range are the most distinguishing. For instance, if subsequences of up to two tokens around a pause produced the most accurate subsequence classifier, transcript-level aggregates should only be extracted for tokens at the D1 and D2 positions in reference to the pause, and not the D3 position.

To validate this method, we create 
five transcript-level feature sets: features aggregated from tokens at the D1 position in reference to a pause (\emph{F-D1}), features aggregated from the D2 position (\emph{F-D2}), features aggregated from the D3 position (\emph{F-D3}), the combination of F-D1, F-D2, and F-D3 (\emph{F-C3}), and the combination of F-D1 and F-D2 (\emph{F-C2}).

\section{Results}
In this section, we report the results for the subsequence and transcript classification experiments.

\subsection{Subsequence Classification}
\label{sec:results-subsequence}
After averaging across four random seeds, M-C2 
was able to achieve an accuracy of 60.7\%, higher than M-C1, M-C3, or M-Utt 
(Tab.\ref{tab:subsquence-results}). 
This leads us to the conclusion that using features from the two tokens preceding and succeeding a pause could  enhance transcript-level classification performance. 

\begin{table}
  \centering
  \begin{adjustbox}{max width=0.8\linewidth}
  \begin{tabular}{ccccc}
    \toprule
    Model & M-C1 & M-C2 & M-C3 & M-Utt \\
    \midrule
     Accuracy & 59.6$\pm$2.6 & \textbf{60.7$\pm$2.5}& 59.8$\pm$0.9 & 60.3$\pm$1.0\\
     \bottomrule
  \end{tabular}
  \end{adjustbox}
  \caption {Subsequence classification performance. Accuracy is averaged across four random seeds.}
  \label{tab:subsquence-results}
\end{table}

\subsection{Transcript Classification}
\label{sec:transcript-results}
We create the F-D1, F-D2, and F-C2 aggregate feature sets, as the highest subsequence classification accuracy was achieved by a model trained on DB-C2. Additionally, in order to validate our claims, we create F-D3 and F-C3.
The highest accuracy of 77.09\% on transcript classification is achieved by an ensemble model that used the \textit{Original} + F-D2 feature set (Tab. \ref{tab:transcript-results}). 

Using one of the four random seeds used to produce the average performance metrics presented in Tab. \ref{tab:transcript-results}, the model using F-D2 features was able to achieve an accuracy of $78.36\%$, the same as the single-seed SOTA accuracy of 78\% \cite{hernandez2018computer}.


\section{Discussion}

As shown in Tab. \ref{tab:subsquence-results} and Tab.~\ref{tab:transcript-results}, features from tokens within 2 tokens of a pause were the most effective in enhancing both subsequence and transcript classification. To determine how these two tasks are connected, we conduct a statistical analysis on the token-level and transcript-level features. 
Two sided t-tests 
between features extracted from tokens found at D1, D2, and D3  
from different classes show similar patterns for features that are significantly different between classes for both the token and transcript-level. Larger concentrations of distinguishing features are found at D1 and D2 than at D3. This could explain the effectiveness of features from the D2 position in both tasks (Tab.~\ref{tab:significance_test}).


However, this pattern congruity does not explain why F-D3 features on their own are more effective than F-D1 features on their own. The trend that both the F-D3 and F-C3 feature sets produced greater transcript-level accuracy than the F-D1 feature set, and lower transcript-level accuracy than the F-D2 and F-C2 feature sets, is the same as the trend for the subsequence classification results reported in Tab. \ref{tab:subsquence-results}. This indicates that subsequence classification may be able to provide better insight into potential transcript classification performance than traditional statistical testing.

It is important to consider the implications of producing a model with the F-D2 feature set that achieved significantly higher accuracy than the most accurate model produced with the F-C3 feature set. As described in Section \ref{sec:transcript-classification}, we perform feature selection using ANOVA F-values for each of the aggregate feature sets. Since F-D2 is a subset of F-C3, this implies that this more traditional feature selection method did not select a group of features from F-C3 that was more effective than the features from F-D2, even though it was able to select any and all of the features in F-D2. This serves as a testament to our feature engineering method, as it demonstrates that even popular feature selection methods are not able to completely remove the negative effects of engineering an excessive amount of ineffective features.

Following several other works that used the DB data set \cite{hernandez2018computer, pou2018learning, sarawgi2020multimodal}, all of our experiments are conducted with K-fold cross validation. While the small size of the DB data set helps to justify this as a validation procedure, optimizing a cross validated performance metric (accuracy, F1, etc.) may lead results using K-fold cross validation to be an overestimate of generalization performance.

DB-C2 produced a more accurate subsequence classifier than any other data subset of DB.
This suggests that the class distinguishing signal from the pause is strongest within a two token radius around the pause. Beyond that radius, the signal may be obstructed by noise from other patterns in speech. However, in different data sets, a different subsequence length may present the strongest, least noisy signal. 
New aggregate features should be created for tokens within whichever range produces the best subsequence classification performance.

However, our results do indicate that there is a strong link between how well features from certain token positions contribute to both subsequence and transcript classification. This may relate to the effect of noise on those token positions, 
which we use subsequence classification to identify.

\begin{table}
  \centering
  \begin{adjustbox}{max width=0.6\linewidth}
  \begin{tabular}{ccc}
    \toprule
   \textbf{Distance} & \textbf{Token-Level} & \textbf{Transcript-Level} \\
    \hline \hline
    D1 & 18 & 12\\
    D2 & 21 & 12\\
    D3 & 7  & 6\\
    \bottomrule
  \end{tabular}
  \end{adjustbox}
  \caption{Number of features that are significantly different between classes according to two sided t-tests for each distance.}
  \label{tab:significance_test}
\end{table}

\section{Conclusion and Future Work}

In this work, we present two principle contributions. First, we describe a novel method for speech classification - subsequence classification - in which speech is modelled as a token of interest, such as a pause, along with surrounding tokens of context. Secondly, we demonstrate how subsequence classification can be used to engineer features that extract distinguishing information while minimizing added noise, and consequently match SOTA performance on a standard data set of CI speech.

Future work should be done to understand why certain context lengths are more conducive for subsequence classification than others, and when that performance can transfer to effective transcript-level classification. Finally, additional work should be done to develop techniques for finding tokens of interest, such as pauses, that can be exploited using our feature engineering technique. 

\clearpage

\bibliography{emnlp2020}
\bibliographystyle{acl_natbib}

\clearpage
\appendix 
\section{Feature list}\label{sec:app-feature-list}
\begin{table*}
  \centering
  \begin{adjustbox}{max width=0.8\linewidth}
  \begin{tabular}{lll}
    \toprule
    Feature    & \# of features & Description \\
    \hline \hline
     Word length & 2 &Length of the word, both in syllables and letters. \\
     Sentiment & 6 & Three measures of the type and intensity of reaction a word produces.\cite{Warriner2013}  \\
     Concreteness & 2 & Measure of the degree to which a word refers to a perceptible entity. \cite{brysbaert2012corpus} \\
     Imageability & 2 & How easy it is for a word to elicit a mental image. \cite{Stadthagen-Gonzalez2006} \\
     Age of acquisition & 2 & Average age that the word is learned. \cite{kuperman2012corpus} \\
     Frequency & 2 & Word counts in a corpus of over 385 million words. \cite{davies2009corpus} \\
     Familiarity & 2 & The perceived popularity of a word.\cite{Stadthagen-Gonzalez2006} \\
     Part of Speech & 5 & Grammatical category of the word.\\
    \bottomrule
  \end{tabular}
  \end{adjustbox}
    \caption{Token-level linguistic features used as input to the models during subsequence classification and their description.}
    \label{tab:features}
\end{table*}
\subsection{Token-Level features}
\label{sec:app-token-features}
For each token, we extract each of the features listed in Tab. \ref{tab:features}. To ensure that each pause has at least one token preceding and succeeding it, \textit{start} and \textit{end} tokens are added to each utterance in DB (Sec. \ref{sec:DB}). For tokens that did not have these features, such as pauses and start/end tokens, all of the features were given a value of zero with the exception of the part of speech. For each feature, with the exception of word length and part of speech, we also extracted the same feature value from the lemmatized token. These features, along with a 5-dimensional, randomly initialized embedding for the part-of-speech, make up the 23-dimensions of each token input vector. Part-of-speech tagging is performed using Spacy \cite{spacy2}. All word tokens then have missing values imputed with feature means, and are then normalized with respect to the feature means and standard deviations of the word tokens (i.e. excluding pauses, start/end tokens).

Identical subsequences found in both classes were removed from each data subset. Additionally, if multiple identical subsequences were found in only one of the classes, all but one of them are removed. Furthermore, we do not include utterances that include only a single pause as the final token, as we assume that this pause is occurring between consecutive sentences, rather than within a single sentence.

\subsection{Transcript-Level Features}
We classify transcripts using 500+ extracted features based on previous literature \cite{fraser2016linguistic, toth2018speech}, which we refer to as the \textit{Original} feature set. Each of these features come from one of 8 categories:
\begin{itemize}
    \item \textbf{Information Units:} Semantic measures, pertaining to the ability to describe concepts and objects in the picture.
    \item \textbf{Discourse Mapping:} Features that help identify cohesion in speech using a visual representation of message organization in speech. We represent each word as a node to build a ‘speech graph’ \cite{mota2012speech}, for the whole transcript. Examples of features extracted include number of edges in this graph, number of  self-loops etc.
    \item \textbf{Coherence:} Semantic continuity that listeners perceive between utterances (locally or globally).
    \item \textbf{Lexical Complexity and Richness :} Different measures of lexical qualities and  variation. Examples of features include  average age of acquisition, number of occurrence of various POS tags etc.
    \item \textbf{Sentiment:} Sentiment lexical norms from Warriner \emph{et al.} \citeyear{Warriner2013}. Examples include average sentiment valence over verbs, average sentiment dominance over nouns etc. 
    \item \textbf{Syntactic Complexity:} Different measures to analyze the syntactic complexity  of speech including features such as number of occurrence of various production rules, mean length of clause (in words) etc.
    \item \textbf{Word finding Difficulty:} Features quantifying difficulty in finding the right words. These include various pause features such as number of filled pauses, pause word ratio etc.
    \item \textbf{Acoustic:} Voice markers such as MFCC coefficients and Zero Crossing Rate (ZCR) related features.
\end{itemize}

All transcript-level features, including the aggregates described in Sec. \ref{sec:app-transcript-aggregates}, have missing values imputed with feature medians. Features are then then standardized by removing the feature median, and scaled according to the range between the first and third quartile \cite{pedregosa2011scikit}.  

\subsection{Transcript-Level Aggregates}
\label{sec:app-transcript-aggregates}

In Sec. \ref{sec:our-technique}, we extend the \textit{Original} set of features with transcript-level aggregates of token-level features. These include averages for each of the features described in Sec. \ref{sec:app-token-features}, with the exception of part of speech. Pauses are not considered when calculating the mean feature values. Additionally, for each part of speech (POS), we include the total amount of times that POS occurs at a certain distance from the pause, divided by the total number of pauses in the transcript multiplied by the percent of words in the transcript that are that POS.

\begin{table*}
  \centering
  \begin{adjustbox}{max width=0.9\linewidth}
  \begin{tabular}{cccccccccc}
    \toprule
   \textbf{Model} & \textbf{Bidirectional} & \textbf{\# of Layers} &
   \textbf{Dropout}&
   \textbf{Epochs}&
   \textbf{Learning Rate}&
   \textbf{Momentum}&
   \textbf{$\lambda$}&
   \textbf{Batch Size} &
   \textbf{Approx. Time}\\
    \hline \hline
    M-C1 & False(12) & 2(10, 5) & False & 600 & 0.01 & 0.9 & 0.0001 & 20 & 6 Min.\\
    M-C2 & True(12) & 2(10, 5)& True(p=0.5)& 600 & 0.01 & 0.9 & 0.0001 & 20 & 17 Min.\\
    M-C3 &  False(50) & 1(40) & True(p=0.5)& 600 & 0.01 & 0.9 & 0.0001 & 20 & 27 Min.\\
    M-Utt & False(50)  & 2(40, 20) & True(p=0.5)& 600 & 0.01 & 0.9 & 0.0001 & 20 & 90 Min.\\
    \bottomrule
  \end{tabular}
  \end{adjustbox}
  \caption{Hyperparameters used by the best performing models for each data subset in the subsequence classification task, as well as the approximate time required to complete cross validation. The number of hidden units in the GRU is indicated in the ``Bidirectional" column, and the number of hidden units in each layer of the predicting network is indicated in the ``\# of Layers" column.}
  \label{tab:subsequence_parameters}
\end{table*}

\begin{table}
  \centering
  \begin{adjustbox}{max width=0.7\linewidth}
  \begin{tabular}{cccc}
    \toprule
   \textbf{Feature Set} & \textbf{Model} & \textbf{\# of Features Selected} &
   \textbf{SMOTE}\\
    \hline \hline
    Original & Ens & - & False\\
    Original w/ feat.sel & Ens & 85 & False\\
    Original + F-D1 & NN & 11& False\\
    Original + F-D2 &  Ens & 15 & False\\
    Original + F-D3 & Ens & 5& True\\
    Original + F-C2 & NN & 20& False\\
    Original + F-C3 & Ens & 5& True\\
    \bottomrule
  \end{tabular}
  \end{adjustbox}
  \caption{Parameters used by the best performing models for each feature set in the transcript classification task.}
  \label{tab:transcript_parameters}
\end{table}

\section{Classification}
\label{sec:appendix-models}
\subsection{Subsequence Classification}
In this work, we perform five-fold cross validation with each of the subsequence data subsets on several different GRU based models \cite{cho2014learning} with attention \cite{yang2016attention}. Each model consists of a GRU that takes the subsequences as input, and outputs to a feed forward neural network which then makes predictions. The attention mechanism uses a linear layer that has as many hidden units as the GRU, as well as a context vector that has as many dimensions as the GRU has hidden units. An extensive search was conducted in terms of finding the most effective model parameters for each data subset. Each model was tested with variations on the number of intermediate layers in the predicting feed-forward network (1, 2 or 3), the addition of dropout, whether the GRU was bidirectional, and the number of hidden units in each layer (\textit{large} or \textit{small}, where \textit{large} has approximately 4 times as many hidden units in each layer as \textit{small}). This creates a total of 24 trials per data-subset. All the models were created with Pytorch~\cite{paszke2017pytorch}, and each model was trained for $600$ epochs using SGD as an optimizer, learning rate $= 0.01$, momentum $= 0.9$, L2 regularization with $\lambda = 0.0001$, batch size of 20, a Cosine Annealing learning rate scheduler,  and cross entropy loss. Each layer with the exception of the final layer uses the ReLU activation function. Additionally, training scripts were run using the CPU of a p2.xlarge Elastic Compute Cloud instance provided by Amazon Web Services\footnote{https://aws.amazon.com/ec2/}.

A summary of the hyperparameters used for each of the best performing models reported in Sec. \ref{sec:results-subsequence} are provided in Tab. \ref{tab:subsequence_parameters}. The number of hidden units in the GRU is indicated in the column indicating whether the GRU was bidirectional, and the number of hidden units in each layer of the predicting network is indicated next to the column indicating the number of layers. 

While selecting the best performing model for DB-C1, DB-C2, DB-C3, and DB-Utt, a model was only considered if it was able to meet or exceed the specificity achieved by a model that was once SOTA, 28.8\%~\cite{di2019enriching}. 

\subsection{Transcript Classification}
We use a Random Forest (100 trees), Gradient Boosting Estimator (with 150 estimators), SVM (with RBF kernel), a 2-layer neural network (NN, 10 units, Adam optimizer, 200 epochs with learning rate initialized to 0.01), and an ensemble of all 4 aforementioned classifiers (Ens) \cite{pedregosa2011scikit}. For each extending feature set described in Sec. \ref{sec:transcript-features}, we jointly optimize the number of features selected from the extending feature set (feature selection with k=3, 5, 7, 9, 11, 13, 15, 20, 25, 30, and all features), whether or not we use the oversampling method SMOTE\cite{chawla2002smote}, and the model type used. For feature selection on the \textit{Original} feature set, we attempted feature selection with k=20, 25, 30, 35, 40, 45, 50, 55, 60, 65, 70, 75, 80, 85, 90, 95, 100, 150, 200, 250, 300, and 350. The best performing configuration for each extending feature set is recorded in Tab. \ref{tab:transcript_parameters}.

\end{document}